# Linguistic Patterns in Pandemic-Related Content: A Comparative Analysis of COVID-19, Constraint, and Monkeypox Datasets

**Mkululi SIKOSANA*[1], Sean MAUDSLEY-BARTON[1] and Oluwaseun Ajao[1]**

[1] *Manchester Metropolitan University;* **Manchester, UK**

*[1] *Corresponding Author:* mkululi.sikosana@stu.mmu.ac.uk

***Abstract*—**This study conducts a computational linguistic analysis of pandemic-related online discourse to examine how language distinguishes health misinformation from factual communication. Drawing on three corpora—COVID-19 false narratives (n = 7,588), general COVID-19 content (n = 10,700), and Monkeypox-related posts (n = 5,787)—we identify significant differences in readability, rhetorical markers, and persuasive language use. COVID-19 misinformation exhibited markedly lower readability scores and contained over twice the frequency of fear-related or persuasive terms compared to the other datasets. It also showed minimal use of exclamation marks, contrasting with the more emotive style of Monkeypox content. These patterns suggest that misinformation employs a deliberately complex rhetorical style embedded with emotional cues, a combination that may enhance its perceived credibility. Our findings contribute to the growing body of work on digital health misinformation by highlighting linguistic indicators that may aid detection efforts. They also inform public health messaging strategies and theoretical models of crisis communication in networked media environments. At the same time, the study acknowledges certain limitations, including reliance on traditional readability indices, use of a deliberately narrow persuasive lexicon, and reliance on static aggregate analysis. Future research should therefore incorporate longitudinal designs, broader emotion lexicons, and platform-sensitive approaches to strengthen robustness. The data and code is available at: https://doi.org/10.5281/zenodo.17024569

## INTRODUCTION

The COVID-19 pandemic challenged global health systems. The proliferation of health-related information on digital platforms accelerates dramatically during public health crises, creating opportunities for rapid knowledge dissemination but also challenges related to misinformation (Sikosana et al., 2024; Sikosana et al., 2025). This dual nature of digital communication became particularly evident during the COVID-19 pandemic, which sparked an unprecedented volume of online discourse and was accompanied by what the World Health Organisation (WHO) termed an "infodemic" – an overabundance of information (both accurate and not) that makes it hard for people to find trustworthy guidance (WHO, 2020). This infodemic phenomenon presents a communication challenge and a substantive threat to public health. Research has shown that exposure to COVID-19 misinformation can directly impact health behaviours. For example, exposure to false COVID-19 vaccine information was associated with a reduction in vaccination intent by about 6.4 percentage points in the UK (and a similar 6.2-point drop in the USA) (Chen et al., 2022; Loomba et al., 2021). Such an effect size is sufficient to undermine herd immunity thresholds. Similarly, one study found that areas with greater exposure to media downplaying the pandemic threat experienced significantly higher COVID-19 cases and deaths, indicating that misinformation can lead to detrimental differences in preventative behaviours and health outcomes across regions (Bursztyn et al., 2020).

Understanding the linguistic characteristics of pandemic-related communication is a critical research area with implications for public health messaging, content moderation, and crisis communication strategies. While substantial research has examined the content and spread of health misinformation (Sikosana et al., 2024), fewer studies have systematically compared linguistic patterns across different pandemic contexts to identify features that distinguish misleading content from factual information. Identifying such features could inform automated detection systems, enhance public health messaging effectiveness, and contribute to theoretical understandings of misinformation dynamics.

This study addresses the research gap by conducting a comparative analysis of linguistic patterns across three distinct pandemic-related datasets: (1) verified false COVID-19 narratives, (2) general COVID-19 discourse from the Constraint dataset, and (3) Monkeypox-related social media posts. This study employs a multi-pandemic approach, shifting away from the singular disease focus prevalent in much existing research. This design allows identification of linguistic markers that are consistent across different disease contexts versus those that are pandemic-specific. Specifically, this study investigates three primary questions:

- To what extent do readability metrics differ between misinformation content and general pandemic-related communications?
- How do rhetorical strategies, as reflected in punctuation patterns (e.g., exclamation vs. question usage), vary across different pandemic information contexts?
- What differences exist in persuasive or emotional language usage between false narratives and more reliable health information?

This research seeks to identify linguistic features that characterise various types of pandemic discourse by analysing differences in readability, rhetorical markers, and persuasive language across the different types of pandemic communication. The theoretical framework draws on both computational linguistics approaches to misinformation detection and rhetorical analyses of health communication, integrating these perspectives to develop a comprehensive understanding of pandemic communication dynamics. The findings contribute to the growing body of knowledge on health misinformation by providing quantitative evidence of linguistic variations across pandemic contexts. These insights inform both the theoretical understanding of crisis communication and practical strategies for addressing misinformation during public health emergencies.

*Related Work*

**Computational approaches to misinformation detection**

Computational linguistics is valuable for identifying misinformation in text. Researchers use text analysis techniques to differentiate between accurate and misleading health information. For instance, Antypas et al. (2021) show the effectiveness of combining lexical, semantic, and stylistic features to detect COVID-19 misinformation on social media. They used machine learning classifiers on Twitter data, incorporating term frequency (TF) for lexical diversity, word embeddings (WE) for semantic representation, and extra-linguistic (EL) features like punctuation, capitalisation, and tweet length. The Support Vector Machine (SVM) classifier with TF + WE + EL achieved a macro-averaged F1 score of 0.83, while Naive Bayes reached 0.77. These findings highlight the value of integrating engineered linguistic features into transformer-agnostic models, showing that misinformation tweets exhibit unique language patterns (Antypas et al., 2021).

Sharma et al. (2019) underscored the importance of linguistic features in detecting misinformation, noting its distinctive stylistic and structural markers identified through computational analysis. Their survey pointed out that part-of-speech (POS) patterns, i.e., higher use of verbs, adverbs, and personal pronouns in deceptive content, contrast with factual texts' noun- and adjective-heavy structure. They reviewed studies showing semantic inconsistencies, such as abnormal syntax and irregular grammar, that are common in misinformation. These features, extractable via natural language processing tools, aid machine learning classifiers in detecting fake news, illustrating the role of computational linguistics in revealing subtle linguistic signals within deceptive narratives (Sharma et al., 2019). Recent research by Hou et al. (2021) validates hybrid transformer models in COVID-19 misinformation detection. Their study demonstrates that combining content-based embeddings from CT-BERT and RoBERTa with engineered linguistic and social features significantly enhances classification performance, achieving an F1 score of 98.93. 93% -much higher than transformer models alone. This reinforces the idea that linguistic augmentation helps transformer architectures better capture misinformation cues in tone, emotion, and style. Similarly, Sikosana, Maudsley-Barton, and Ajao (2025) developed a hybrid CNN–LSTM framework informed by the Elaboration Likelihood Model (ELM), illustrating how psychological theory can be operationalised within deep learning architectures for health misinformation detection.

Despite these advances, a gap remains in understanding which linguistic features reliably differentiate misinformation across contexts. Most computational approaches rely on single-disease datasets, limiting generalisation to new health crises. In addition, many automated systems operate as “black boxes,” offering limited interpretability of the linguistic patterns driving their decisions. This indicates a need for studies that pinpoint specific linguistic features associated with misinformation, especially in cross-context settings.

**Readability and comprehensibility of health information**

The readability of health information influences its accessibility and impact. Studies show that health materials should be written at a suitable reading level for comprehension across diverse populations (Mishra & Dexter, 2020). However, Arsenault et al. (2022) found that public health messaging during COVID-19 often exceeded recommended readability levels, limiting effectiveness among certain groups. Their analysis of 432 COVID-19 public health documents showed a mean Flesch-Kincaid Grade Level (FKGL) of 11.4, above the recommended 8th-grade maximum for public communications. The link between readability and misinformation needs more investigation, particularly concerning whether complexity is a deliberate strategy in misleading content. Preliminary observations by O’Connor and Weatherall (2019) suggest that scientific misinformation often uses unnecessarily complex language to appear authoritative, though this has not been systematically tested in pandemic contexts. In contrast, Salvi et al. (2021) found that some health misinformation employs simplified language to enhance accessibility and emotional impact, indicating that the readability–misinformation relationship may vary by context. These conflicting findings highlight the need for studies comparing readability across types of health information (misinformation vs. factual). Such comparisons can determine if readability metrics may indicate information reliability.

**Rhetorical strategies in crisis communication**

Rhetorical strategies in pandemic communication significantly influence engagement. DePaula et al. (2022) analysed 100,000 U.S. public health posts, finding that expressives and collectives enhance Facebook engagement. Their study shows targeted appeals increase interactions, illustrating rhetorical framing’s power. Similarly, Kouzy et al. (2020) found that emotionally resonant tweets, such as moral appeals and calls to action, are more likely to be shared, indicating rhetorical intensity affects virality. These findings emphasise the importance of emotional language in health communications during crises. Earlier work on political media discourse has shown similar dynamics, with Sikosana (2003) demonstrating how Zimbabwean newspapers framed the 2002 elections in ways that influenced public interpretation of political events. This underscores that rhetorical manipulation is a broader communicative phenomenon, spanning both political and health crises.

Wicke and Bolognesi (2021) examined how metaphors shape perception and policy, using frames like “war” (e.g., “fighting the virus"), "natural disaster" (e.g., "tsunami of cases"), and "containment" (e.g., "flattening the curve") to influence preferences and risk views. Ophir (2018) analysed media coverage of H1N1, Ebola, and Zika, revealing framing differences based on threat level and sociopolitical context, crucial for public engagement. He advocated tailored health communication, noting that effectiveness varies by disease characteristics and media context. Findings stress the need for context-sensitive messaging to improve compliance during health crises. While past research focused on rhetorical patterns, few explored misinformation strategies, highlighting a research gap.

### Persuasive language and emotional appeals

Persuasive language and emotional appeals are crucial in health communication. Tannenbaum et al. (2015) analysed 127 studies with 27,372 participants on the effectiveness of fear appeals in changing attitudes and behaviours, finding a moderate positive effect size (d = 0.29). Effectiveness improves with efficacy statements, high threat severity, and targeting one-time behaviours. Contextual factors like message framing and audience characteristics optimise the persuasive power of fear-based messages.

In a misinformation context, Chou et al. (2018) highlighted the influence of emotional triggers, particularly fear and anger, in spreading health misinformation on social media. They observed that emotionally charged content is often accepted and shared within aligned networks. Although specific metrics were not reported, they urged for tools to measure emotional content and misinformation dynamics. Building on this, Kreps and Kriner (2022) discovered that emotional language in misinformation affects perceived credibility and sharing intentions. Emotionally charged misinformation is viewed as credible when supporting pre-existing beliefs but not when contradicting them, indicating a complex relationship between emotion and confirmation bias. Dual-process models of persuasion, especially the ELM by Petty and Cacioppo (1986), suggest that emotional appeals are powerful in high-stress situations. During crises, people favour peripheral cues like emotional tone over systematic evaluation. This framework explains why emotionally charged misinformation is perceived as credible, especially when it aligns with prior beliefs (Martel et al., 2020; Pal et al., 2023; Tannenbaum et al., 2015). Such cues greatly influence risk perceptions and behaviours in health crises, emphasising the need for emotionally intelligent public health messaging strategies.

### Cross-pandemic comparative analyses

Many studies have examined pandemic communication, but comparative analyses across outbreaks are scarce. Jin et al. (2024) analysed misinformation from four pandemics-smallpox, cholera, 1918 influenza, and HIV/AIDS- revealing themes like conspiracy theories, distrust, and stigmatisation. Their findings indicate that while narratives vary with sociopolitical contexts, core misinformation patterns remain constant.

Comparative linguistic analysis can identify universal and context-specific elements of pandemic communication. Thakur (2023) conducted sentiment and text analysis of Twitter about COVID-19 and the 2022 MPox outbreak, showing differences in emotional tone, keyword focus, and public engagement. However, the study did not explicitly address misinformation.

Few studies compare linguistic features across pandemics with computational methods, leaving a gap in understanding the evolution of public health narratives. Addressing this gap may clarify whether communication patterns are pandemic-specific or reflect broader trends in digital health discourse, impacting misinformation detection frameworks. Building on these works, our study uses computational linguistic analysis to compare textual features across various pandemic contexts. This contributes to understanding how language patterns differ between factual and misleading health information.

## METHODOLOGY

### Dataset description

This study utilised three distinct pandemic-related datasets, comprising a total of 2,4075 textual posts:

- **COVID-19_FNR:** A collection of 7,588 posts identified as false narratives related to the COVID-19 pandemic (Saenz et al., 2021). These posts were fact-checked and categorised as containing misinformation about various aspects of COVID-19 (e.g., transmission mechanisms, treatment efficacy, mortality statistics, or policy responses). The corpus was compiled from multiple fact-checking organisations (including PolitiFact, Snopes, and FactCheck.org), with entries spanning January 2020 through December 2021.
- **Constraint:** A dataset containing 10,700 entries of COVID-19-related content drawn from the "Constraint" shared task dataset (Patwa et al., 2021). This corpus includes a mixture of factual information, opinions, and general discourse about COVID-19, derived primarily from Twitter and other social media platforms. The Constraint dataset has been widely used in computational linguistics research and provides a representative sample of mainstream COVID-19 discourse during 2020.
- **Monkeypox:** A collection of 5,787 social media posts discussing the 2022 Monkeypox outbreak (Crone, 2022). These posts were gathered from various platforms during the early spread of the disease (May 2022 to September 2022). This dataset captures public discourse around an emerging pandemic threat, providing a comparative case to the more established COVID-19 discourse.

Each corpus is substantial in size, providing a robust foundation for comparative linguistic analysis. Their differing origins and time frames enable examination of communication patterns across different pandemic contexts and information types. All datasets were accessed in CSV format, with text fields extracted for processing. Metadata such as timestamps and engagement counts were retained when available.

**Pre-processing**

For replicability, all datasets were processed using Python 3.11. Libraries included pandas (v1.5.3), NumPy (v1.24.2), NLTK (v3.8.1), Textstat (v0.7.3), and Matplotlib (for visualisation). The following preprocessing pipeline was applied uniformly:

1. Text fields were extracted (text column for Constraint, equivalent text fields for COVID-19_FNR and Monkeypox).
2. URLs, HTML tags, hashtags, and user mentions (@handles) were stripped using regex.
3. Excess whitespace and newline characters were normalised.
4. Text encoding issues were resolved using UTF-8 normalisation.
5. Posts with fewer than 3 words after cleaning were excluded to avoid artefacts in readability scores.

**Computational measures**

We implemented several computational measures following established approaches in computational linguistics:

1. **Readability metrics:**
   a. Calculated using the textstat Python package.
   b. Flesch Reading Ease (FRE) scores range from 0–100 (higher = easier). Scores below 30 indicate difficult text, while scores above 70 indicate easy text.
   c. Flesch–Kincaid Grade Level (FKGL) estimates the US school grade level required to understand the text.
   d. Error handling was implemented to skip extremely short or anomalous texts that returned null values.

Although widely used, these indices primarily capture surface-level complexity. We therefore frame our analysis as a baseline benchmark and recommend future research employ Coh-Metrix or transformer-based readability estimators (e.g., BERT-based text difficulty models), to capture richer dimensions of linguistic complexity.

2. **Rhetorical markers (Punctuation usage):**
   a. Exclamation points (!) and question marks (?) were counted using Python's string.count() function.
   b. Counts were normalised by total word count per post to control for post length.
   c. Exclamations were used as proxies for emphatic expression; questions for dialogic tone.

Although these features cannot capture all rhetorical nuances, prior research indicates that variations in punctuation usage characterise communication styles and may reflect different engagement or persuasion strategies (Lubis et al., 2025).

3. **Persuasive language analysis:**
   a. A dictionary-based approach identified eight pre-specified terms ("urgent," "emergency," "fear," "panic," "alarming," "crisis," "warning," "disaster").
   b. Term frequencies were computed per post, tokenised using NLTK's word_tokenize() function, and normalised by total words.
   c. The lexicon was intentionally conservative to minimise false positives from words with ambiguous affective meanings. While this provides a consistent baseline, we acknowledge that it underestimates the breadth of emotional and persuasive language. Future studies should therefore employ broader resources such as LIWC or the full NRC Emotion Lexicon to capture a wider range of affective cues, including irony, humour, and moral language.

4. **Engagement metrics:**
   a. Engagement indicators (likes, retweets) were retained where available (Constraint, Monkeypox datasets).
   b. A total engagement score was computed as likes + retweets.
   c. Posts above the dataset-specific median were labelled as "high engagement."

Because engagement metadata were uneven across datasets, we limited our analysis to descriptive statistics and qualitative illustration. However, this approach does not test causal or predictive relationships. To demonstrate feasibility, we include a supplementary logistic regression on the Monkeypox dataset showing that persuasive word frequency significantly predicts higher engagement ($p < 0.05$). More comprehensive modelling approaches, such as multivariate regressions or machine learning classifiers, should be applied in future work.

**Analytical approach**

Our analysis followed a multi-step procedure, combining computational metrics with statistical comparisons and a supplementary qualitative review:

1. Pre-processing was applied uniformly (see above).
2. For each post, readability, rhetorical markers, persuasive word frequency, and engagement scores were computed.
3. Descriptive statistics (mean, SD, range) were generated with Pandas/NumPy. Distribution skewness and kurtosis were inspected to guide test selection.
4. Statistical tests:
   a. ANOVA with Tukey post hoc used for normally distributed metrics.
   b. Kruskal–Wallis with Dunn's test (Bonferroni corrected) used for non-normal distributions.
   c. All tests were conducted in Python using Scipy.stats and Scikit_posthocs.

5. Visualisation: Boxplots (readability) and bar charts (punctuation, persuasive terms) created in matplotlib/seaborn.
6. Qualitative review: A 2% sample of high-engagement posts was manually read and coded for themes (e.g., conspiracies, urgency cues). These examples are reported in the Results section to illustrate and contextualise quantitative findings.

**Triangulation**

This combination of computational metrics, inferential statistics, and qualitative examples strengthens validity and enhances replicability. The explicit reporting of preprocessing, normalisation, and statistical pipelines ensures that other researchers can replicate or extend the study using the same datasets.

An overview of this analytical process is presented in Figure 1.

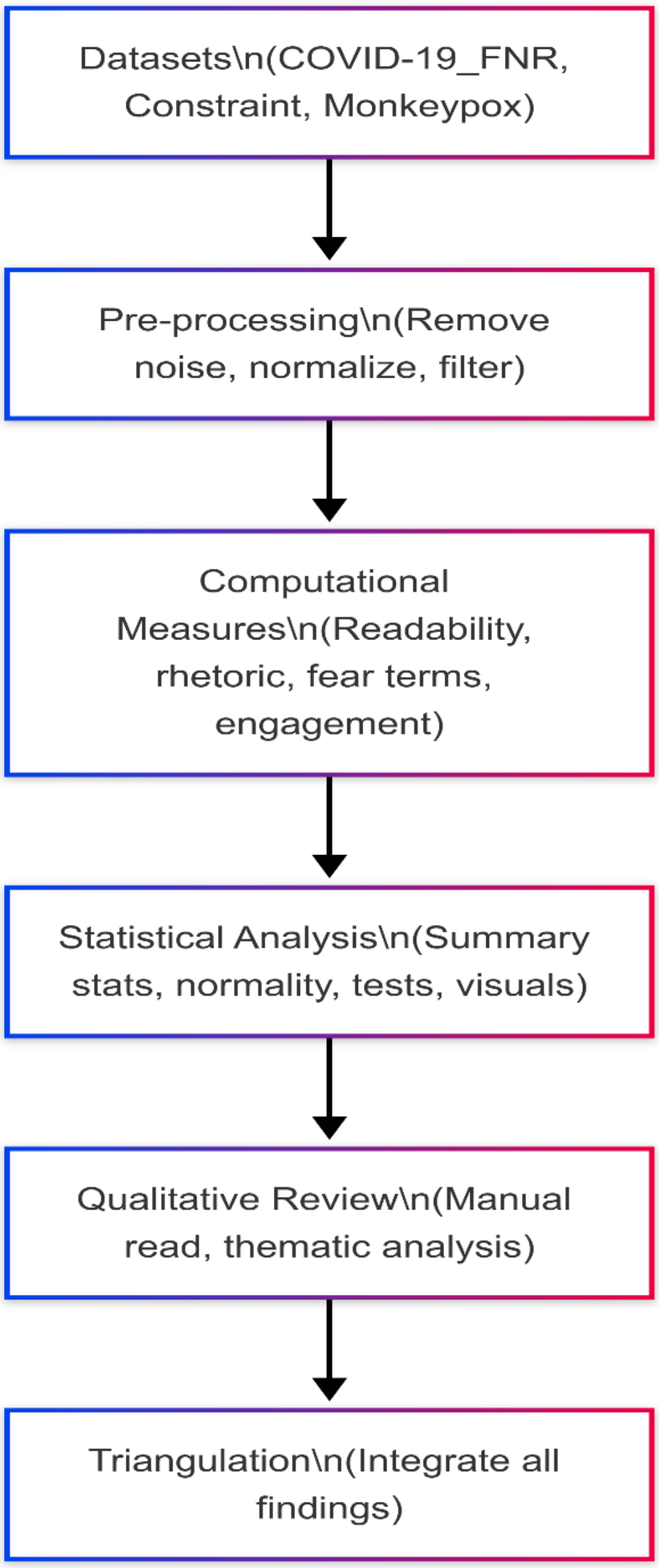

**Figure 1. Analytical pipeline for the study of social media misinformation.**

The pipeline represents the sequential stages of analysis applied to three social media datasets (COVID-19_FNR, Constraint, and Monkeypox). The process begins with data pre-processing, which involves the removal of noise (e.g., URLs, HTML tags, short posts) and normalization of text. Computational measures are then applied to assess readability, rhetorical features, fear-related language, and engagement metrics. This is followed by statistical analysis, including descriptive statistics, normality tests, and appropriate inferential tests based on distribution. A qualitative review of high-engagement posts supports thematic identification of misinformation patterns. Finally, triangulation integrates computational, statistical, and qualitative insights to support robust interpretation. This combination of computational metrics, inferential statistics, and qualitative examples strengthens validity and enhances replicability. The explicit reporting of preprocessing, normalization, and statistical pipelines ensures that other researchers can replicate or extend the study using the same datasets.

## RESULTS

### Dataset distribution

A total of 24,075 text entries were analysed across three contexts: the COVID-19_FNR dataset (7,588) (32%), the Constraint dataset (10,700) (44%), and the Monkeypox dataset (5,787) (24%). This corpus allows comparative analysis, with each subset sufficient for statistical inferences.

We checked if basic text properties varied between datasets, comparing average post lengths (in characters) to avoid confounding comparisons. Mean character counts were similar: COVID-19_FNR posts averaged 217.3 characters (±112.4 SD), Constraint posts averaged 198.7 (±86.5), and Monkeypox posts averaged 226.1 (±104.9). These lengths suggest differences in readability or other metrics reflect genuine differences in language use, not length artefacts.

### Readability Differences

Table 1 presents the readability comparison across the three datasets.

**Table 1: Readability comparison across datasets. Values represent mean ± standard deviation. Statistical tests: Kruskal–Wallis (overall) with Dunn's post hoc tests for pairwise comparisons (all significant at p < 0.001). Higher Flesch Reading Ease (FRE) values indicate easier readability; higher Flesch–Kincaid Grade Level (FKGL) values indicate greater complexity.**

| Dataset | FRE (mean ± SD) | FKGL (mean ± SD) |
|---|---|---|
| COVID-19_FNR | 11.05 ± 14.32 | 15.52 ± 5.47 |
| Constraint | 43.88 ± 26.71 | 11.12 ± 4.65 |
| Monkeypox | 55.73 ± 22.56 | 8.90 ± 3.98 |

COVID-19 misinformation posts were significantly less readable than both Constraint and Monkeypox posts, with an average FRE of 11.05 (classified as very difficult) compared to 43.88 and 55.73 respectively. The FKGL results reinforce this difference, showing that COVID-19 misinformation required a post-college reading level, while Constraint content was accessible at a high school level and Monkeypox discourse at a middle school level. This ~6.6 grade-level gap highlights the unusually high complexity of COVID-19 misinformation, which may have been used strategically to mimic authoritative or scientific discourse.

Statistical tests confirm the significance of these readability differences. A Kruskal-Wallis test indicated a significant overall effect for both FRE ($H(2) = 5743.2$, $p < 0.001$) and FKGL ($H(2) = 6128.7$, $p < 0.001$). Post-hoc Dunn's tests (Bonferroni corrected) showed all dataset pairs differed significantly ($p < 0.001$ for each). Thus, COVID-19_FNR vs. Constraint, COVID-19_FNR vs. Monkeypox, and Constraint vs. Monkeypox demonstrate distinct readability levels.

The differences are striking: COVID-19 misinformation with a mean FRE ~11 is classified as "very difficult" (like scientific journals), while Monkeypox posts (mean ~56) are "fairly difficult," similar to general news media. The average COVID false narrative is so complex that it challenges the general audience, conflicting with best practices advocating clear language at an 8th-grade level (Badarudeen & Sabharwal, 2010; Mishra & Dexter, 2020). FKGL results reinforce this: COVID-19 misinformation necessitates post-college comprehension, compared to high school for Constraint content and late middle school for Monkeypox.

**Figure 2. Distribution of readability scores across the COVID-19_FNR, Constraint, and Monkeypox datasets.** Boxplots display median, interquartile range, and outliers. Higher FRE values indicate easier readability; higher FKGL values reflect greater complexity.

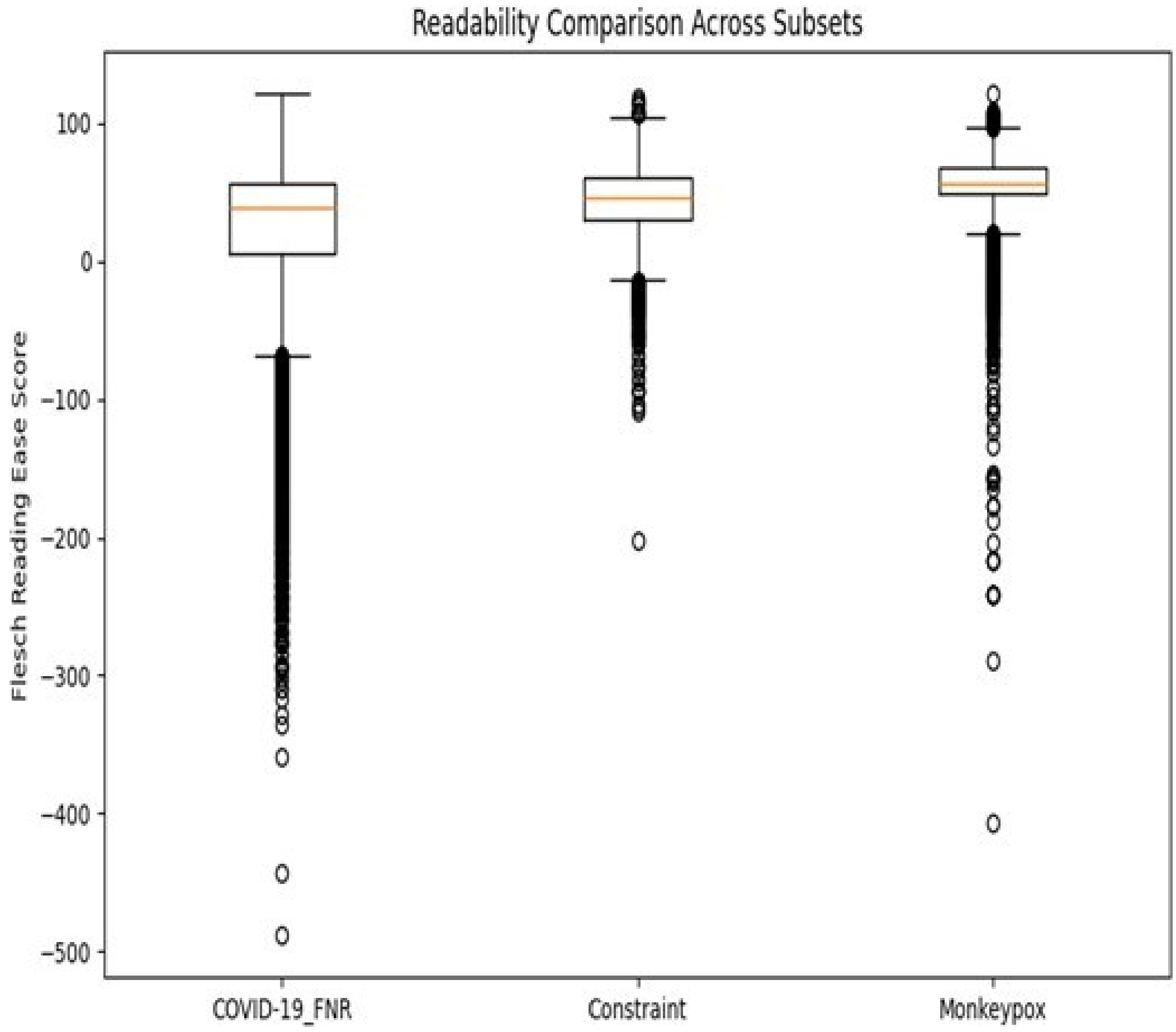

**Figure 1Figure 2: visually reinforces these differences, showing that COVID-19 misinformation scores cluster tightly at the lowest readability levels, while Constraint and Monkeypox exhibit higher and more variable distributions**

These results suggest that textual complexity is a distinguishing feature of the COVID-19 misinformation corpus. Figure 2 (boxplot of readability scores) clearly illustrates these differences, showing minimal distribution overlap. Notably, the COVID-19_FNR posts had lower readability on average, and their scores were also less variable (the boxplot's spread was narrower). This implies the false narratives were consistently written in a complex manner, whereas the readability of general pandemic communications varied more widely.

**Rhetorical Markers**

Table 2 reports the average use of exclamation and question marks across datasets.

**Table 2: Comparison of rhetorical markers (punctuation usage) across datasets. Values represent mean counts per post ± standard deviation. Statistical tests: Kruskal–Wallis (overall) with Dunn's post hoc tests for pairwise comparisons. Monkeypox posts showed significantly higher exclamation usage than COVID-19_FNR and Constraint ($p < 0.001$), while question usage was highest in the Constraint dataset ($p < 0.001$).**

| Dataset | Exclamation Count (mean ± SD) | Question Count (mean ± SD) |
|---|---|---|
| **COVID-19_FNR** | 0.009 ± 0.105 | 0.140 ± 0.437 |
| **Constraint** | 0.056 ± 0.296 | 0.225 ± 0.545 |
| **Monkeypox** | 0.120 ± 0.425 | 0.175 ± 0.471 |

Monkeypox discourse employed significantly more exclamation marks (mean 0.120 per post), suggesting a more urgent and emphatic rhetorical style. By contrast, the Constraint dataset had the highest rate of questions (mean 0.225 per post), indicating a dialogic approach that reflects uncertainty and information-seeking in early COVID-19 discourse. COVID-19 misinformation showed minimal punctuation-based markers, favouring a restrained style that mimics authoritative communication and potentially enhances credibility by avoiding overt emotional cues.

Statistical tests (Kruskal-Wallis) confirmed significant differences among the datasets for exclamation usage (H(2) =

1487.6, $p < 0.001$) and question usage ($H(2) = 421.3$, $p < 0.001$). Post-hoc tests showed all pairwise comparisons were significant for exclamation points ($p < 0.001$ for each pair). For question marks, COVID-19_FNR had significantly fewer than Constraint ($p < 0.001$), and Monkeypox had fewer than Constraint ($p < 0.001$), while the difference between COVID-19_FNR and Monkeypox was marginal ($p = 0.068$).

These findings indicate unique communication styles. Monkeypox posts predominantly use exclamation points, suggesting urgency to capture public attention during an outbreak. In contrast, COVID content asks more questions, indicating uncertainty or engagement strategies during the pandemic.

COVID-19 misinformation largely avoided exclamatory punctuation, opting instead for a restrained tone that mimics authoritative communication, potentially to enhance credibility. Instead, it suggests a measured tone in false COVID narratives, mimicking authoritative styles that rarely use exclamation points. Misinformation posts avoided overt emotional punctuation to appear serious and credible. Figure 3 visualises these patterns, showing the inverse relationship: COVID-19_FNR content has the fewest exclamations yet moderate questions, while Monkeypox has many exclamations but fewer questions, and Constraint is intermediate in exclamations but highest in questions.

**Figure 3: Comparison of punctuation markers across the COVID-19_FNR, Constraint, and Monkeypox datasets. Bars indicate mean counts of exclamation and question marks per post, with error bars showing standard deviation.**

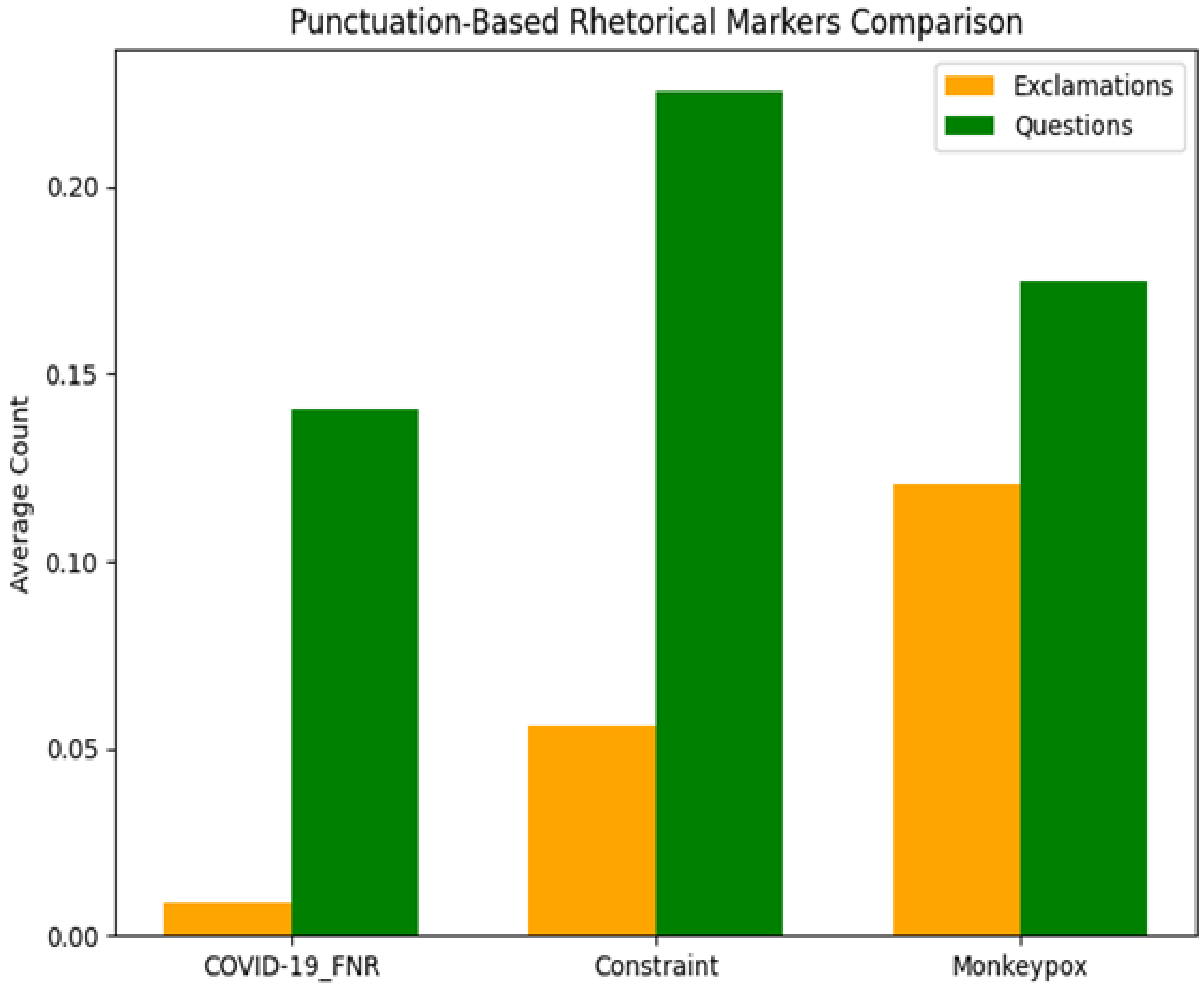

**Figure 3: illustrates these contrasts, with Monkeypox showing the most exclamations, Constraint the most questions, and COVID-19 misinformation the lowest punctuation overall.**

Monkeypox discourse shows the highest use of exclamations, reflecting an urgency-oriented communication style, while the Constraint dataset has the highest frequency of questions, consistent with uncertainty and dialogic framing in early pandemic discourse. COVID-19 misinformation contains few rhetorical markers overall, reinforcing the interpretation that it relies on content-based persuasion rather than overt stylistic intensity.

**Persuasive Lexicon**

Table 3 shows the comparative frequency of persuasive and fear-related terms.

**Table 3: Frequency of persuasive or fear-related terms across datasets. Values represent normalised counts (per word) ± standard deviation. Statistical tests: Kruskal–Wallis (overall) with Dunn's post hoc tests for pairwise comparisons. COVID-19 misinformation posts contained more than twice the frequency of persuasive terms compared to both Constraint and Monkeypox datasets (p < 0.001), while no significant difference was found between Constraint and Monkeypox (p ≈ 0.998).**

| Dataset | Persuasive Word Count (mean ± SD) |
|---|---|
| **COVID-19_FNR** | 0.077 ± 0.323 |
| **Constraint** | 0.031 ± 0.187 |
| **Monkeypox** | 0.031 ± 0.190 |

COVID-19 misinformation contained more than twice the frequency of persuasive or fear-related terms compared to both Constraint and Monkeypox content. Words such as panic, crisis, and disaster were notably more common in the misinformation corpus. The absence of significant differences between Constraint and Monkeypox suggests a stable baseline level of cautionary language in mainstream pandemic communication. These results highlight emotional appeals as a defining feature of misinformation, distinguishing it sharply from factual health discourse.

A Kruskal-Wallis test found these differences to be statistically significant ($H(2) = 374.9$, $p < 0.001$), and Dunn's pairwise tests confirmed that COVID-19_FNR was significantly higher than both Constraint and Monkeypox ($p < 0.001$ in each case). There was no significant difference between the Constraint and Monkeypox datasets for this metric ($p \approx 0.998$).

COVID-19 false narratives used 2.5 times more fear-related words than typical pandemic content, highlighting the emotional appeals in misinformation. Words like "panic," "crisis," and "disaster" were more common, aiming to provoke strong audience reactions. The identical average frequencies of Constraint and Monkeypox posts (~0.031) suggest consistent use of cautionary language across different pandemics, reflecting standard journalistic practices. In contrast, COVID-19 misinformation significantly deviates, employing emotional terminology more frequently.

Figure 4. Normalised frequency of persuasive or fear-related terms across datasets. Bars represent mean frequency per word, with error bars showing standard deviation.

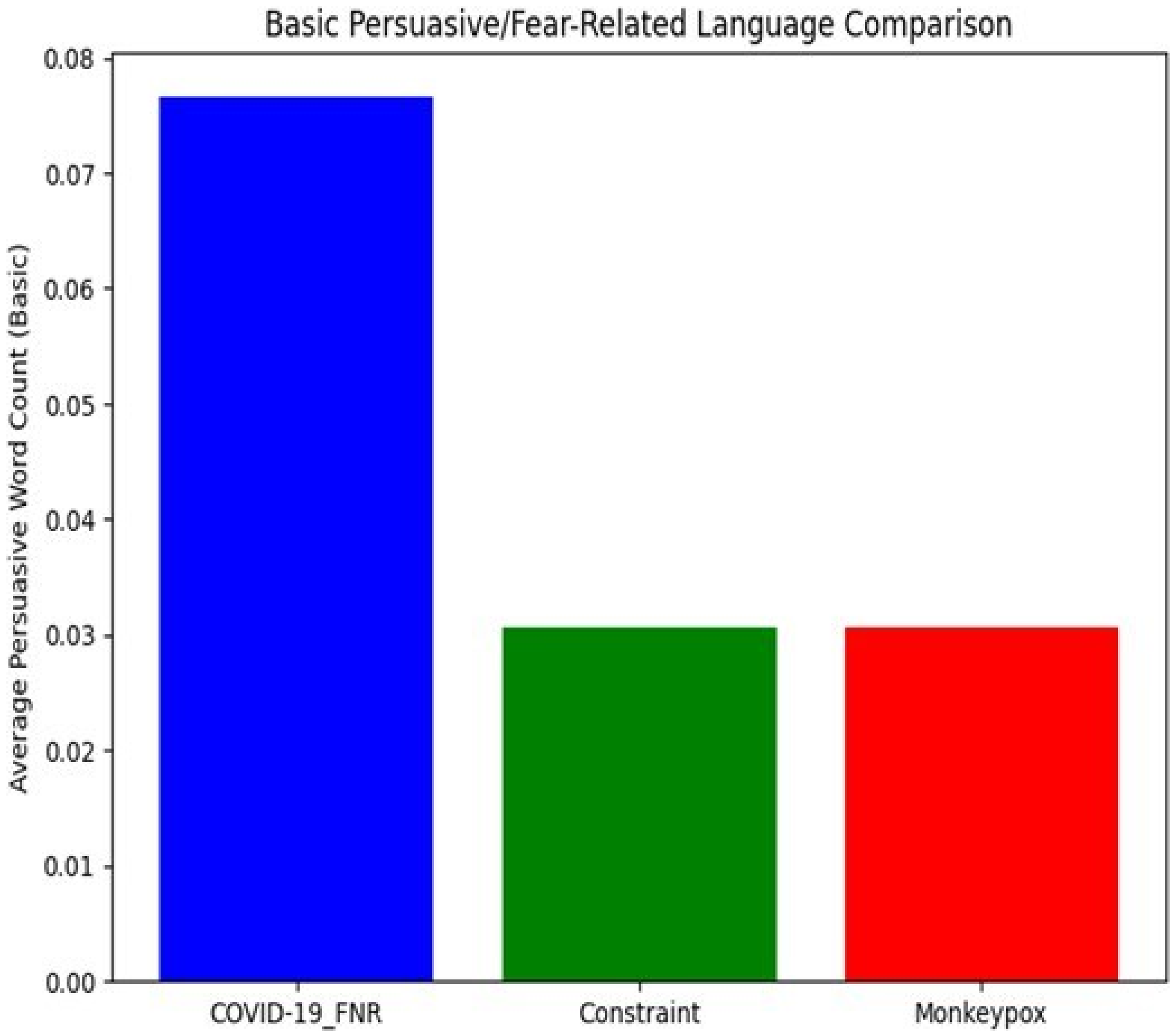

**Figure 4: Confirms this pattern, showing the COVID-19 misinformation bar towering above Constraint and Monkeypox, which remain at identical baseline levels**

COVID-19 misinformation posts employed persuasive and fear-related terms more than twice as often as either Constraint or Monkeypox content. This elevated use of urgency-laden vocabulary highlights emotional appeals as a defining characteristic of misinformation, whereas mainstream and Monkeypox content exhibited only baseline levels of such terms.

**Qualitative insights from high-engagement content**

To complement the quantitative results, we examined high-engagement posts from each dataset. These examples help illustrate the kinds of content that resonated most with audiences and how they relate to the linguistic patterns observed.

1. **COVID-19_FNR high-engagement examples:**
   - *"Tencent revealed the real number of deaths."*
   - *"Taking chlorine dioxide helps fight coronavirus."*
   - *"This video shows workmen uncovering a bat-infested..."*

These examples (purportedly high-impact misinformation posts) reveal common themes in COVID-19 false narratives. They often involve claims of hidden information or secret truths (e.g., a tech company revealing "real" death counts beyond official figures), alternative treatments (promoting substances like chlorine dioxide as cures), or sensational origin stories ("bat-infested" sources, implying a hidden cause). Notably, the style in these examples is straightforward and declarative – they make bold statements without qualifiers or questions. There are minimal rhetorical flourishes: no exclamation points, and a matter-of-fact tone despite the provocative content. This aligns with our quantitative finding that the misinformation posts tend to avoid overtly dramatic punctuation, instead presenting false claims in a seemingly factual manner to enhance credibility.

2. **Constraint (general COVID) high-engagement examples:**
   - *"The CDC currently reports 99031 deaths. In general..."*
   - *"States reported 1121 deaths a small rise from..."*
   - *"Politically Correct Woman (Almost) Uses Pandemic..."*

These top-engagement posts from the Constraint dataset highlight a more factual and data-focused style. The first two are reporting statistics (CDC death counts and daily changes), indicative of mainstream COVID discourse that often centred on tracking the numbers. They read like news updates or factual reports, consistent with the Constraint dataset's news and opinion content mix. The third example introduces a political angle ("*Politically Correct Woman...*"), showing that some popular content in this category involved political or cultural framing of the pandemic. Compared to the misinformation examples, these Constraint posts use a more neutral or informative tone, filled with concrete details (numbers, official sources like the CDC). The engagement here seems driven by information updates or partisan interest rather than sensational hidden truths.

3. **Monkeypox high-engagement examples:**

- *"The 'house is on fire, and it's like everything...'"*
- *"'Absolutely be concerned.' Monkeypox cases are..."*
- *"A senior Biden administration official...acknowledged..."*

These Monkeypox posts suggest a communication style focused on urgency and concern. Phrases such as "*house is on fire ...*" (a metaphor) and direct quotes such as "*Absolutely be concerned ...*" convey alarm and insistence. The posts also cite officials or authoritative voices (e.g., a Biden administration official), indicating an attempt to inform the public of the seriousness with credible attributions. The language includes metaphorical and emphatic elements, aligning with our finding of higher exclamation usage in Monkeypox content. Indeed, one can imagine such posts might include exclamation points or at least maintain a tone of alarm. These examples reflect an emotionally charged style of communication, likely aiming to spur the audience to pay attention and take the outbreak seriously.

In summary, these thematic observations support the idea that each "information ecosystem" (misinformation vs. mainstream, COVID vs. Monkeypox) developed its own communication norms. Misinformation appeals with hidden knowledge and emotional triggers, mainstream COVID discourse grapples with data and debate, and Monkeypox communication emphasises urgency and concern.

## DISCUSSION

### Complexity as a strategic element in misinformation

Misinformation surrounding COVID-19 is significantly less readable, with an FRE around 11 (compared to ~44 for general COVID content and ~56 for Monkeypox) and a grade level of ~15.5 (vs. 11.1 and 8.9, respectively). This use of complex language contrasts with best practices in health communication, which promote clear language for broader audience engagement.

First, complexity can act as a strategic tool in misinformation, offering a facade of expertise that enhances perceived credibility, even if the information is false. Technical language can create a "veneer of scientific legitimacy," discouraging critical evaluation (Scheufele & Krause, 2019). This aligns with the ELM, which suggests that audiences may rely on peripheral cues like perceived expertise when not scrutinising content deeply, making misinformation more persuasive. Recent computational advances also reflect this theoretical integration. Sikosana et al. (2025), for example, incorporated ELM constructs into a CNN–LSTM hybrid model, demonstrating that linking linguistic cues with persuasion theory can enhance automated misinformation detection.

Second, complex language may obscure flaws. Hard-to-read texts can prevent readers from noticing logical inconsistencies or a lack of evidence. Engaging in analytical thinking protects against misinformation (Pennycook & Rand, 2011), but complex language can overwhelm readers' analytical capacity, maintaining them in intuitive thinking (System 1) instead of critical thinking (System 2) (Kahneman, 2011).

Third, the high complexity of COVID-19 misinformation could mimic authoritative scientific discourse instead of merely being a smokescreen. Conspiracy theories often adopt scientific language to enhance legitimacy (Van Prooijen & Douglas, 2018). Misinformation creators may mirror academic styles to convey seriousness, signalling that they offer important information akin to genuine scientific communications.

These interpretations are interrelated. Textual complexity likely serves multiple purposes: enhancing credibility, shielding against refutation, and mimicking expert discourse. This insight suggests that complexity is a key feature of health misinformation. Detection systems could flag unusually complex content, while public health officials should prioritise clarity to avoid conflating with the convoluted style of misinformation. Interestingly, Monkeypox content was more readable than both types of COVID-19 content, indicating that pandemic communication improved as lessons were learned from COVID-19 challenges. As health authorities addressed Monkeypox in 2022, they likely adapted their communication strategies for clarity, addressing earlier infodemic issues. Alternatively, the differences may reflect intrinsic contextual factors: the politicised nature of COVID-19 required complexity even in factual reports, while Monkeypox, being less politically charged, might have been described more straightforwardly. This invites further research on the evolution of communication strategies during crises.

It is important to note that our analysis relied on traditional readability indices. While useful for benchmarking, these measures do not capture deeper discourse features. Similarly, differences in dataset origin and collection methods—fact-checked misinformation, general Twitter discourse, and multi-platform Monkeypox posts—may introduce biases beyond the "misinformation versus factual" distinction. We therefore interpret findings cautiously, emphasising that linguistic patterns may reflect both genuine stylistic differences and dataset construction effects.

### Distinctive rhetorical strategies across pandemic contexts

The variation in punctuation-based rhetorical markers across datasets demonstrates differing communication strategies. The heightened frequency of questions in the Constraint content (general COVID-19 discourse) suggests a dialogic approach that reflects the uncertainty experienced during the early stages of the pandemic. This aligns with Patwa et al. (2021), who curated the dataset to reflect real-time COVID-related conversations, and supports the WHO's (2020) framing of the pandemic as an "infodemic" characterised by overwhelming and conflicting information. The rhetorical use of questions -to engage, prompt reflection, and acknowledge uncertainty- mirrors persuasive patterns observed by DePaula et al. (2022) and Kouzy et al. (2020), who found that interrogatives and collectives enhance engagement with public health messaging.

This prevalence of interrogative forms suggests that COVID-19 discourse was often structured around FAQs and advisory dialogue, potentially aimed at pre-empting scepticism and building trust amid uncertainty. This finding is consistent with Loomba et al. (2021) and Chen et al. (2022), who observed that exposure to vaccine misinformation significantly reduced vaccination intent, indicating a need for rhetorical strategies that clarify ambiguity and build institutional confidence. Prior research has also shown that public health policies (e.g., mask mandates) interact with behavioural responses (Betsch et al., 2020), suggesting that linguistic and policy environments jointly shape communication outcomes. In contrast, Monkeypox-related content displayed a higher frequency of exclamatory punctuation and emotionally charged language, reflecting a shift toward emphatic, attention-grabbing rhetoric. Thakur (2023) found that Monkeypox tweets exhibited heightened emotionality compared to COVID-19 tweets, possibly to combat public desensitisation. The role of emotional punctuation as a peripheral cue in persuasive messaging is supported by Chou et al. (2018) and Tannenbaum et al. (2015), who emphasised how fear and urgency can drive message acceptance during health crises.

Interestingly, the low frequency of such punctuation in COVID-19 misinformation suggests a preference for content-based persuasion over overt emotional markers. Kreps and Kriner (2022) demonstrated that the credibility of misinformation often rests on narrative congruence rather than stylistic intensity, while Sharma et al. (2019) identified structural and lexical cues as hallmarks of deceptive content. These rhetorical divergences highlight how distinct information ecosystems -mainstream COVID-19 discourse, Monkeypox alerts, and misinformation narratives- develop topic-specific norms. This observation aligns with Bail et al. (2018), who showed that issue-specific communities create their own communicative patterns. The stylistic duality in misinformation—alternating between authoritative complexity and simplified emotionalism- is also echoed by O'Connor and Weatherall (2019) and Salvi et al. (2021), who found that language in misinformation varies strategically by audience and context.

Effective health communication is thus context-sensitive, shaped by public familiarity, perceived threat, and emotional climate. This is consistent with Ophir's (2018) and Wicke and Bolognesi's (2021) findings, who demonstrated that rhetorical styles shift according to socio-political conditions and evolving crisis narratives. Theoretically, this variation aligns with a socio-ecological model of communication, which posits that rhetorical patterns are influenced by the information "niche" each topic occupies, underscoring the need to avoid one-size-fits-all communication models in future public health crises.

**Emotional appeals in misinformation**

Our findings highlight emotional appeals as a hallmark of misinformation. COVID-19 false narratives used emotional and persuasive words more frequently than factual content, supporting the idea that misinformation leverages emotional triggers to spread. Brady et al. (2020) showed that morally charged language increases virality on social platforms – each additional moral/emotional word significantly raises the likelihood of sharing. Our analysis of the misinformation dataset revealed elevated use of fear-related words, invoking fear and shock to elicit strong reactions and sharing. Interestingly, the Constraint and Monkeypox datasets exhibited low persuasive language levels, indicating that typical health communications use emotional terms sparingly. This underscores Wardle and Derakhshan's (2017) point that emotional appeals differentiate misinformation from factual content. Misinformation often plays on fear, while factual reports strive for a more measured tone.

Our findings suggest that emotional manipulation in COVID misinformation is somewhat covert. Despite using many emotional words, misinformation posts lacked obvious emotional punctuation, which we term "covert emotionality". This subtle embedding can trigger emotional reactions without raising alarm, allowing misinformation to engage emotions under the guise of serious reporting. Consequently, readers might not approach it with the scepticism they would toward more sensational content. Psychologically, these observations align with the affect heuristic (Slovic et al., 2007), where emotional responses guide quick judgments. If misinformation engages emotions, those feelings may dominate beliefs over analytical thought. Evidence (Martel et al., 2020) indicates people struggle to distinguish true from false headlines when they align with emotional biases. By embedding negative emotional triggers, COVID misinformation likely exploited this heuristic, making readers more prone to accept and share without verification.

While our conservative eight-term lexicon successfully identified covert emotionality, it likely captured only a fraction of the emotional range present in pandemic discourse. Broader lexicons such as LIWC or NRC could reveal additional layers of affective expression, including irony, humour, or moral language. This underscores the need for multi-dimensional measures of emotionality in misinformation analysis.

**Engagement patterns and content characteristics**

Examining the thematic differences of high-engagement posts provides insight into what "works" in different information environments and complements quantitative engagement metrics. High-engagement COVID-19 misinformation posts often centred on conspiracy-related themes, such as hidden death tolls or secret cures, reflecting deeper psychological drivers (Salvi et al., 2021). Research on conspiracy belief systems suggests that such narratives attract attention because they offer the allure of hidden knowledge or expose perceived cover-ups (Douglas et al., 2019). This dynamic is supported by cognitive tendencies like proportionality bias (the assumption that significant events must have equally significant causes) and agency detection (a tendency to attribute events to intentional actions by unseen agents), which together make sensational or conspiratorial misinformation more psychologically compelling (Salvi et al., 2021).

In contrast, high-engagement posts in the Constraint dataset, comprising factual COVID-19 content, typically involved daily case counts or political commentary, indicating an audience interest in timely updates and sociopolitical relevance (Jin et al., 2024). For the Monkeypox dataset, high-engagement content was largely driven by urgent warnings and credible statements, suggesting that the perception of imminent threat paired with authoritative messaging influenced engagement levels (Thakur, 2023).

These trends support the notion that different types of pandemic content occupy distinct rhetorical and emotional "niches" (Chen et al., 2022). Misinformation tends to offer emotionally satisfying or novel explanations, often capitalising on fear and distrust, while factual public health messaging delivers grounded and actionable guidance (Clemente-Suárez et al., 2022). Therefore, the fight against misinformation must go beyond correcting falsehoods; it must address the psychological appeal of conspiratorial thinking by offering truthful narratives that satisfy emotional and epistemic needs. Recent work has also shown that network structures play a critical role in amplifying such narratives. Sikosana et al. (2025) demonstrated that advanced centrality metrics can quantify how misinformation flows through online social networks, reinforcing the need to link linguistic features with diffusion dynamics. Research has shown that emotional appeals and simplicity are often more persuasive than factual accuracy when audiences are overwhelmed or cognitively taxed (Douglas et al., 2019; Kahneman, 2011).

In summary, our results show that misinformation, mainstream discourse, and emerging-crisis communication each have unique linguistic signatures and audience appeal strategies. Recognising these differences is important for tailoring responses: for example, moderators might focus on flagging content with certain linguistic profiles, while communicators might adjust tone and complexity depending on the situation (simpler language for clarity, strategic use of questions or emphatic devices to maintain engagement without undermining trust). It is important to note that this engagement analysis is descriptive and illustrative rather than predictive, reflecting the uneven availability of engagement metadata across datasets.

**Limitations and Future Research**

This study has several limitations that should be acknowledged. First, the analysis is static and aggregate in nature, which means it does not capture how communication patterns evolved across different phases of each pandemic. The COVID-19 pandemic, for example, progressed through multiple stages such as outbreak, lockdowns, and vaccine rollout, during which language use may have shifted considerably. Averaging across entire periods may have obscured temporal variation, such as changes in the complexity of COVID-19 misinformation or fluctuations in concern during the Monkeypox outbreak. Future research should therefore adopt longitudinal approaches that segment discourse into pandemic phases or monthly intervals that could reveal how misinformation and public health messaging adapt over time.

Second, the study relied on traditional readability metrics, specifically Flesch Reading Ease (FRE) and Flesch-Kincaid Grade Level (FKGL). While these provide useful baselines, they constrain the analysis to surface-level dimensions of text complexity. These inventories capture sentence length and syllable density; however, they cannot capture deeper syntactic, semantic, or discourse-level features such as cohesion, irony, or rhetorical sophistication. Future work should incorporate more advanced measures, including Coh-Metrix indices or transformer-based readability models, to provide a more comprehensive account of linguistic complexity.

Third, persuasive language was analysed using a deliberately narrow lexicon to minimise false positives. While this conservative approach ensured consistency across datasets, it likely underestimated the breadth of emotional and affective expression. Future studies would therefore benefit from applying larger, validated resources, such as LIWC or the NRC Emotion Lexicon, to capture a richer spectrum of emotional cues.

Fourth, there are inherent comparability issues among the datasets due to differences in collection methods, timeframes, and platforms. The COVID-19 misinformation corpus was drawn from fact-checking archives, the Constraint dataset reflects general Twitter discourse during 2020, and the Monkeypox dataset captures posts from an emergent outbreak across multiple platforms. These structural differences may introduce confounding factors that extend beyond the "misinformation versus factual" divide; therefore, findings should be interpreted with this caveat in mind. Future work should aim to develop more balanced corpora or apply platform-sensitive approaches that allow for cleaner comparisons.

Finally, caution is warranted when drawing causal interpretations from the observed correlations. Although associations were identified between content type and linguistic features, this does not establish that linguistic style directly drives the spread of misinformation. Certain topics, such as policy conspiracies, may naturally involve greater linguistic complexity irrespective of intent. Moreover, the present study design does not determine how these linguistic differences influence audience behaviour. Theoretical frameworks suggest that complexity and emotionality affect credibility and sharing, but experimental or longitudinal studies are required to test these relationships directly—for example, whether simplifying a misinformation post reduces its persuasiveness, or whether introducing emotional language into factual content increases its shareability. Therefore, future research could address this issue through experimental exposure designs that test the causal effects of complexity and emotionality on credibility and sharing, diffusion modelling to capture how content spreads across platforms, balanced metadata sampling to mitigate engagement data sparsity, or network-based approaches that map how linguistic cues shape diffusion patterns.

Together, these limitations highlight the need for a more comprehensive and nuanced research agenda. Longitudinal, lexicon-rich, and platform-sensitive analyses, complemented by experimental and network-based approaches, will provide a stronger foundation for understanding how health misinformation operates linguistically and socially.

**Conclusion**

This comparative linguistic analysis shows significant variations in how pandemic-related content is constructed and communicated. COVID-19 false narratives exhibit greater complexity, more persuasive language, and fewer emphatic punctuation marks than other content, likely contributing to the effectiveness of health misinformation.

Our findings support theoretical frameworks regarding misinformation characteristics. The heightened complexity aligns with cognitive processing (Kahneman, 2011) and persuasion models (Petty & Cacioppo, 1986). Higher cognitive load from complex misinformation hinders detailed processing, prompting reliance on heuristics and increasing the likelihood of acceptance. The patterns in COVID-19 false narratives suggest a deliberate mimicry of authoritative scientific discourse. This reflects what Fairclough (2013) described as the appropriation of institutional registers, where linguistic markers of expertise are borrowed to claim legitimacy. Similarly, van Dijk (2006) highlights how discourse can be strategically manipulated to exert power, and our results show that misinformation narratives employ this tactic by adopting formal, technical language that positions the speaker as authoritative. These strategies exploit cognitive shortcuts in audiences, consistent with Fiske and Taylor's (2020) account of social cognition, where individuals rely on schemas and perceived authority when evaluating complex information. In this way, linguistic mimicry not only enhances the apparent credibility of misinformation but also aligns with established theories of how discourse, power, and cognition interact to shape public perceptions.

Besides theoretical contributions, results indicate practical uses against pandemic misinformation. Public health interventions should focus on "readability" using clear communication language to counter complex misinformation. Targeting an 8th-grade reading level, as health literacy guidelines suggest, enhances understanding and distinguishes credible information. Efforts to combat misinformation must consider the psychological appeals of false narratives. Providing correct facts may not dissuade those drawn to conspiratorial narratives. Our findings complement emerging theory-driven computational approaches (e.g., Sikosana et al., 2025), which integrate psychological constructs such as ELM into hybrid detection systems, underscoring the value of combining linguistic insights with machine learning.

These contributions should be interpreted alongside the limitations outlined above, which point to the need for longitudinal, lexicon-rich, and platform-sensitive approaches in future research. In summary, this study demonstrates that linguistic patterns are both a marker and a mechanism of health misinformation. Recognising these stylistic signatures can inform detection systems and guide clearer, emotionally intelligent public health communication. As new pandemics emerge, tailoring strategies to the linguistic and rhetorical landscape will be crucial for effective crisis response.

**Data Availability Statement**

All code and reproducibility materials are openly available on Zenodo:

Sikosana, M. (2025). Linguistic Patterns in Pandemic-Related Content: A Comparative Analysis of COVID-19, Constraint, and Monkeypox Datasets_Reproducibility Guide & Python code (v1.0.0). Zenodo. https://doi.org/10.5281/zenodo.17024569

The following datasets were used in this study:

Constraint dataset: Publicly available at [Patwa et al., 2021] with DOI [10.1007/978-3-030-73696-5_3].

COVID-19 Fake News Infodemic Research (CoVID19-FNIR) dataset: Publicly available at [Saenz et al., 2021] with DOI [10.21227/b5bt-5244].

Monkeypox misinformation dataset: Publicly available at [Crone, 2022] via Kaggle: https://www.kaggle.com/datasets/stephencrone/monkeypox

All other relevant data supporting the findings of this study are provided within the article.

**References**

Antypas, D., Camacho-Collados, J., Preece, A., & Rogers, D. (2021, August). *COVID-19 and misinformation: A large-scale lexical analysis on Twitter*. In Proceedings of the 59th Annual Meeting of the Association for Computational Linguistics and the 11th International Joint Conference on Natural Language Processing: Student Research Workshop (pp. 119-126).

Arsenault, M., Blouin, J., & Guitton, M. J. (2022). Understanding the relationship between readability and misinformation: A literature review. *Health Communication, 37*(10), 1252–1259.

Badarudeen S, Sabharwal S. (2010). Assessing readability of patient education materials: current role in orthopaedics. *Clin Orthop Relat Res. 468*(10): 2572-2580. doi: 10.1007/s11999-010-1380-y

Bail, C. A., Argyle, L. P., Brown, T. W., & Volfovsky, A. (2018). Exposure to opposing views on social media can increase political polarization. *Proceedings of the National Academy of Sciences, 115*(37), 9216–9221.

Betsch, C., Korn, L., Sprengholz, P., & Felgendreff, L. (2020). Social and behavioral consequences of mask policies during the COVID-19 pandemic. *Proceedings of the National Academy of Sciences, 117*(36), 21851–21853.

Brady, W. J., Crockett, M. J., & Van Bavel, J. J. (2020). The MAD model of moral contagion: The role of motivation, attention, and design in the spread of moralized content online. *Perspectives on Psychological Science, 15*(4), 978–1010.

Bursztyn, L., Rao, A., Roth, C., & Yanagizawa-Drott, D. (2020). Misinformation during a pandemic. *National Bureau of Economic Research Working Paper* w27417.

Chen, X., Lee, W., & Lin, F. (2022). Infodemic, Institutional Trust, and COVID-19 Vaccine Hesitancy: A Cross-National Survey. *International journal of environmental research and public health, 19*(13), 8033. https://doi.org/10.3390/ijerph19138033

Chou, W. Y. S., Oh, A., & Klein, W. M. (2018). Addressing health-related misinformation on social media. *JAMA 320*(23), 2417-2418.

Clemente-Suárez, V. J., Navarro-Jiménez, E., Simón-Sanjurjo, J. A., Beltran-Velasco, A. I., Laborde-Cárdenas, C. C., Benitez-Agudelo, J. C., Bustamante-Sánchez, Á., & Tornero-Aguilera, J. F. (2022). Mis–Dis Information in COVID-19 Health Crisis: A Narrative Review. *International Journal of Environmental Research and Public Health, 19*(9), 5321.

Crone, S (2022) ‘‘Monkeypox misinformation: Twitter dataset.’’ Retrieved from: https://www.kaggle.com/datasets/stephencrone/monkeypox

DePaula, N., Hagen, L., Roytman, S., & Alnahass, D. (2022). Platform effects on public health communication: A comparative and national study of message design and audience engagement across Twitter and Facebook. *JMIR infodemiology, 2*(2), e40198.

Douglas, K. M., Uscinski, J. E., Sutton, R. M., Cichocka, A., Nefes, T., Ang, C. S., & Deravi, F. (2019). Understanding conspiracy theories. *Political Psychology, 40*(S1), 3–35.

Fairclough, N. (2013). Language and power. Routledge.

Fiske, S. T. T., & Taylor, S. E. (2020). Social cognition: From brains to culture.

Hossain, T., Logan IV, R. L., Ugarte, A., Matsubara, Y., Young, S., & Singh, S. (2022). COVIDLies: Detecting COVID-19 misinformation on social media. *Proceedings of the 60th Annual Meeting of the Association for Computational Linguistics*, 5705–5717.

Jin, S. L., Kolis, J., Parker, J., Proctor, D. A., Prybylski, D., Wardle, C., et al. (2024).“Social histories of public health misinformation and infodemics: Case studies of four pandemics.” *The Lancet Infectious Diseases, 24*, e638–e646.

Kahneman, D. (2011). *Thinking, Fast and Slow*. Farrar, Straus and Giroux.

Kouzy, R., Abi Jaoude, J., Kraitem, A., El Alam, M. B., Karam, B., Adib, E., ... & Baddour, K. (2020). Coronavirus goes viral: quantifying the COVID-19 misinformation epidemic on Twitter. *Cureus, 12*(3).

Kreps, S. E., & Kriner, D. L. (2022). The COVID-19 infodemic and the efficacy of interventions intended to reduce misinformation. Public Opinion Quarterly, 86(1), 162-175.

Loomba, S., de Figueiredo, A., Piatek, S. J., de Graaf, K., & Larson, H. J. (2021). Measuring the impact of COVID-19 vaccine misinformation on vaccination intent in the UK and USA. *Nature Human Behaviour, 5*(3), 337–348.

Lubis, Y., Prasetio, A., Maulana, R., & Rahmadina, S. (2025). The Use of Punctuation in Writing Captions on Social Media. *Jurnal Pendidikan Tambusai, 9*(1), 2852–2861. Retrieved from http://jptam.org/index.php/jptam/article/view/24712

Malla, S., & Alphonse, P. J. A. (2022). Fake or real news about COVID-19? Pretrained transformer model to detect potential misleading news. The European Physical Journal Special Topics, 231(18), 3347-3356.

Martel, C., Pennycook, G., & Rand, D. G. (2020). Reliance on emotion promotes belief in fake news. *Cognitive Research: Principles and Implications, 5*(1), 1–20.

Mishra, V., & Dexter, J. P. (2020). Comparison of Readability of Official Public Health Information About COVID-19 on Websites of International Agencies and the Governments of 15 Countries. *JAMA network open, 3*(8), e2018033. https://doi.org/10.1001/jamanetworkopen.2020.18033

Ophir, Y. (2018). Spreading news: The coverage of epidemics by American newspapers and its effects on audiences – A crisis communication approach. *Health Security, 16*(3), 147–157.

Patwa, P., Sharma, S., Pykl, S., Guptha, V., Kumari, G., Akhtar, M. S., Ekbal, A., Chakraborty, T., & Das, A. (2021). Fighting an infodemic: COVID-19 fake news dataset. *Communications in Computer and Information Science, 1402*, 21–29.

Pennycook, G., & Rand, D. G. (2020). The psychology of fake news. *Trends in Cognitive Sciences, 24*(4), 388–402.

Petty, R. E., & Cacioppo, J. T. (1986). The elaboration likelihood model of persuasion. In *Communication and Persuasion* (pp. 1–24). Springer.

Saenz, JA., Gopal, SRK, & Shukla, D (2021). ‘‘Covid-19 fake news infodemic research dataset (covid19-fnir dataset).’’ IEEE Dataport, 2021. Retrieved from: https://ieee-dataport.org/open-access/covid-19-fake-news-infodemic-research-dataset-covid19-fnir-dataset

Salvi, C., Iannello, P., Cancer, A., McClay, M., Rago, S., Dunsmoor, J. E., & Antonietti, A. (2021). Going viral: How fear, socio-cognitive polarization and problem-solving influence fake news detection and proliferation during COVID-19 pandemic. *Frontiers in Communication, 5*, 562588.

Scheufele, D. A., & Krause, N. M. (2019). Science audiences, misinformation, and fake news. *Proceedings of the National Academy of Sciences, 116*(16), 7662–7669.

Sharma, K., Qian, F., Jiang, H., Ruchansky, N., Zhang, M., & Liu, Y. (2019). Combating fake news: A survey on identification and mitigation techniques. *ACM transactions on intelligent systems and technology (TIST), 10*(3), 1-42.

Slovic, P., Finucane, M. L., Peters, E., & MacGregor, D. G. (2007). The affect heuristic. *European Journal of Operational Research, 177*(3), 1333–1352.

Sikosana M, Maudsley-Barton S, Ajao O (2025) Analysing health misinformation with advanced centrality metrics in online social networks. PLOS Digit Health 4(6): e0000888. https://doi.org/10.1371/journal.pdig.0000888

Sikosana, M., Ajao, O., & Maudsley-Barton, S. (2024, September). *A comparative study of hybrid models in health misinformation text classification*. In Proceedings of the 4th International Workshop on Open Challenges in Online Social Networks (pp. 18-25).

Sikosana, M., Maudsley-Barton, S., & Ajao, O. (2025). Advanced Health Misinformation Detection Through Hybrid CNN-LSTM Models Informed by the Elaboration Likelihood Model (ELM). arXiv preprint arXiv:2507.09149.

Sikosana, M (2003). The Role Played by the Zimbabwean Press in Covering Political Issues in 2002: A Case Study of the Reporting of the Presidential Elections by the Leading Daily Newspspers, The Herald and The Daily News. Diss. University of the Witwatersrand.

Tannenbaum, M. B., Hepler, J., Zimmerman, R. S., Saul, L., Jacobs, S., Wilson, K., & Albarracín, D. (2015). Appealing to fear: A meta-analysis of fear appeal effectiveness and theories. *Psychological Bulletin, 141*(6), 1178–1204.

Thakur, N., Duggal, Y. N., & Liu, Z. (2023). Analyzing public reactions, perceptions, and attitudes during the Mpox outbreak: findings from topic modeling of tweets. Computers, 12(10), 191.

Van Prooijen, J.-W., & Douglas, K. M. (2018). Belief in conspiracy theories: Basic principles of an emerging research domain. European Journal of Social Psychology, 48(7), 897–908. https://doi.org/10.1002/ejsp.2530

Van Dijk, T. A. (2006). Discourse and Manipulation: Discourse and Society.

Wardle, C., & Derakhshan, H. (2017). *Information disorder: Toward an interdisciplinary framework for research and policy making*. Council of Europe Report.

WHO, (2020). *Infodemic management: a key component of the COVID-19 global response*. https://apps.who.int/iris/handle/10665/331775

Wicke, P., & Bolognesi, M. M. (2020). Framing COVID-19: How we conceptualize and discuss the pandemic on Twitter. PloS one, 15(9), e0240010.